\pdfoutput=1

\documentclass[11pt]{article}

\usepackage[]{acl}
\usepackage[frozencache=true,cachedir=minted-cache]{minted} 

\usepackage{times}
\usepackage{latexsym}
\usepackage{graphicx}
\usepackage{booktabs}
\usepackage[all]{tcolorbox}

\usepackage{makecell}

\usepackage{colortbl}  
\usepackage{array}   
\usepackage{subcaption} 

\usepackage{multirow}
\usepackage{amsmath}  

\usepackage[T1]{fontenc}

\usepackage[utf8]{inputenc}

\usepackage{microtype}
\usepackage{amsfonts}
\usepackage{amssymb}

\usepackage{inconsolata}

\title{Hypertext Entity Extraction in Webpage}

\author{Yifei Yang$^{\spadesuit}$ \quad Tianqiao Liu$^{\clubsuit}$ \quad Bo Shao$^{\clubsuit}$ \quad Hai Zhao$^{\spadesuit}$\thanks{$\ $ Corresponding author.}\\ \bf Linjun Shou$^{\clubsuit}$\footnotemark[1] \quad Ming Gong$^{\clubsuit}$ \quad Daxin Jiang$^{\clubsuit}$\\
$^{\spadesuit}$Shanghai Jiao Tong University \quad $^{\clubsuit}$Microsoft STCA\\
{\tt yifeiyang@sjtu.edu.cn \quad zhaohai@cs.sjtu.edu.cn} \\ \tt \{tianqiaoliu,boshao,lisho,migon,djiang\}@microsoft.com}

\begin{document}
\maketitle
\begin{abstract}
Webpage entity extraction is a fundamental natural language processing task in both research and applications. Nowadays, the majority of webpage entity extraction models are trained on structured datasets which strive to retain textual content and its structure information. However, existing datasets all overlook the rich hypertext features (e.g., font color, font size) which show their effectiveness in previous works. To this end, we first collect a \textbf{H}ypertext \textbf{E}ntity \textbf{E}xtraction \textbf{D}ataset (\textit{HEED}) from the e-commerce domains, scraping both the text and the corresponding explicit hypertext features with high-quality manual entity annotations. Furthermore, we present the \textbf{Mo}E-based \textbf{E}ntity \textbf{E}xtraction \textbf{F}ramework (\textit{MoEEF}), which efficiently integrates multiple features to enhance model performance by Mixture of Experts and outperforms strong baselines, including the state-of-the-art small-scale models and GPT-3.5-turbo. Moreover, the effectiveness of hypertext features in \textit{HEED} and several model components in \textit{MoEEF} are analyzed.
\end{abstract}

\section{Introduction}

Webpage entity extraction is a fundamental and challenging task in natural language processing (NLP) that aims to accurately locate and extract a diverse set of predefined entities from the heterogeneous landscape of web content, such as dates, times and locations \cite{lockard2020zeroshotceres, bhardwaj2021web, cao2021extracting}. It can provide support for a multitude of downstream tasks while also promoting recommendation systems and search engines, assuming a pivotal role.

\begin{figure}[!tp]
    \centering
    \includegraphics[width=0.95\linewidth,scale=1.00]{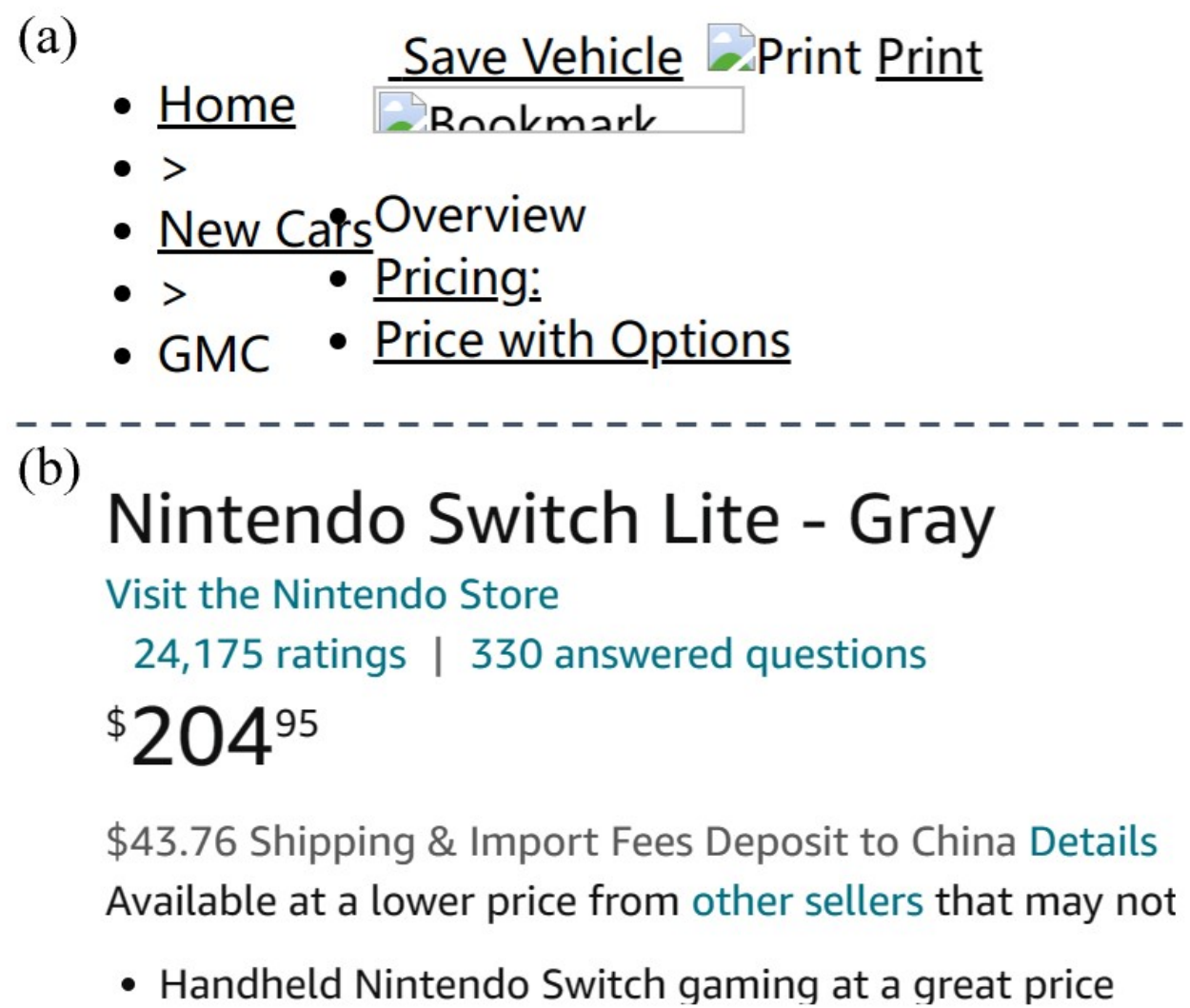}
    \caption{(a) A webpage from SWDE, which lacks hypertext features and contains noise. (b) A webpage from our \textit{HEED} that keeps hypertext information.}
    \label{fig:swde_case}
\end{figure}

Currently, most of the webpage entity extraction models are trained on structured datasets such as SWDE \cite{hao2011one}, expanded SWDE \cite{lockard2019openceres} that preserve the text along with the DOM (Document Object Model) trees\footnote{A DOM tree is a collection of nodes, where each node has its XPath address and original text content \cite{lin2020freedom}.} in the webpages. However, these datasets only strive to retain the entirety of the textual content and its structure information, often neglecting the encompassing hypertext features that webpages possess, such as font color and positions of elements, which have been proven to be highly effective in information extraction \cite{chen2009polyuhk, wong2009learning}. We showcase a rendered webpage from SWDE in Figure \ref{fig:swde_case} (a), which only keeps text and the basic hierarchical structure of the webpage, while rich hypertext information such as font size and color are missing. The overlapping text and expired images in Figure \ref{fig:swde_case} also indicate the presence of significant noise in the dataset. To this end, a high-quality dataset with rich hypertext information and an entity extraction model leveraging both text and hypertext features need to be proposed. Thus, in this paper:

$\bullet$ We collect a unique webpage entity extraction dataset called \textbf{H}ypertext \textbf{E}ntity \textbf{E}xtraction \textbf{D}ataset (\textit{HEED}) which is compiled by harnessing data from search engine shopping service records.\footnote{We will make our dataset public after the paper is accepted.} We carefully select the top domains within this service, including Amazon, eBay, aliexpress, etc. Then we conduct website crawling to extract both text and hypertext features. We gather a comprehensive array of hypertext features by our self-innovated parsing tool to extract various attributes, such as element bounding boxes, font sizes, and a diverse range of other pertinent information. We show a rendered webpage from HEED in Figure \ref{fig:swde_case} (b), which retains rich hypertext information such as font size and color compared to Figure \ref{fig:swde_case} (a). We also annotate the Image, Name, and Price entities with professional annotators.

$\bullet$ We further propose an innovative feature fusion solution for incorporating different features called \textbf{Mo}E-based \textbf{E}ntity \textbf{E}xtraction \textbf{F}ramework (\textit{MoEEF}) based on Mixture of Experts. It significantly enhances model performance without remarkably increasing the complexity of the backbone model and also outperforms strong baselines.

$\bullet$ We conduct detailed ablation studies and in-depth analysis to prove the effectiveness of extracted hypertext features in HEED and several model components in MoEEF.

\section{HEED}

\begin{figure*}[!tp]
    \centering
    \includegraphics[width=0.98\linewidth,scale=1.00]{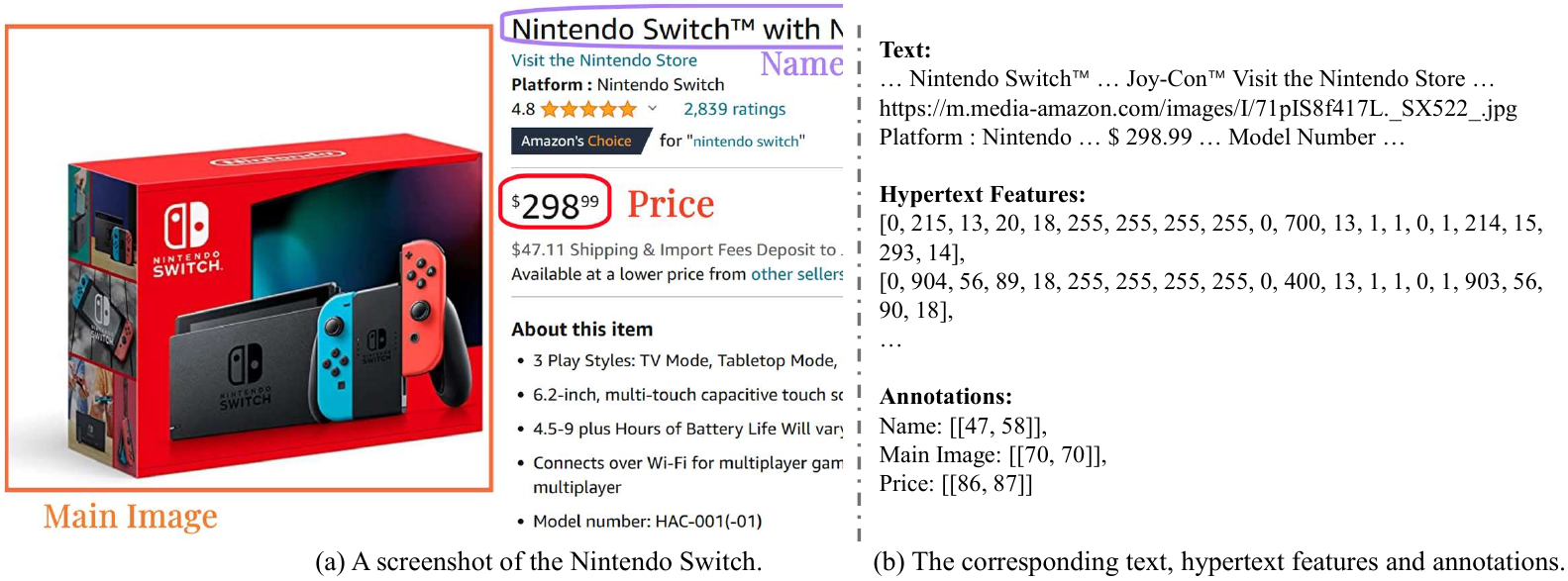}
    \caption{A sample from HEED.}
    \label{fig:main_amazon}
\end{figure*}

HEED is a multi-lingual webpage entity extraction dataset derived from e-commerce domains with rich hypertext features. We present in Figure \ref{fig:main_amazon} a sample from it, showing the original webpage as well as the extracted features and annotated entities.

\subsection{Data Sources}

In our search engine records, there are numerous e-commerce pages sourced from well-known global platforms such as Amazon, eBay, aliexpress, etc. We consider different platforms as distinct domains. The languages from these domains exhibit a high degree of diversity, including English (En), French (Fr), Chinese (Zh) and so on. We collect the data from Apr 2022 to Oct 2022 and have performed data anonymization. The dataset sizes and corresponding numbers of domains from some popular languages are in Appendix \ref{app:dataset_details} Table \ref{table:static}.

\subsection{Features Extraction}

\subsubsection{Text}
To avoid obtaining heavily tedious DOM tree data like in previous work, we directly extract plain text from webpages by crawling the text content in accordance with the left-to-right and top-to-bottom order on the webpage. We skip irrelevant elements such as the search box and input field. When encountering images and logos, we store their hyperlinks. Note that these hyperlinks are discarded in some previous datasets, so HEED includes more comprehensive information. The text crawled from each webpage is stored in a single line of plain text, with adjacent tokens separated by a single space.

\subsubsection{Hypertext Features}\label{sec:vis feat}
Inspired by the fact that humans perceive webpages heavily relying on visual cues such as the position, size and color of the text provided by the hypertext features, we extract the features associated with the text. Specifically, every token has its unique features, which can be classified into 5 categories:

\textbf{Font-Style Features.} As font size and font weight directly reflect the importance of text in a webpage, with larger and bolder text typically serving as emphasis, we extract \textbf{\textit{font-size}} and \textbf{\textit{font-weight}}. Furthermore, since the \textbf{\textit{font-color}} can explicitly represent the price change, product category, etc., we also extract it, which is represented by the RGBA four-channel format.

\textbf{Bounding Box Features.} In the rendered webpage, each element has a bounding box. These bounding boxes can locate elements and reveal webpage structure. We separately extract the \textbf{\textit{element bounding box}} and \textbf{\textit{token bounding box}}. Typically, continuous elements with the same CSS (Cascading Style Sheet) are wrapped within the same element bounding box, resulting in the same element bounding box feature. The token bounding box wraps each token separately, providing more granular position and structural information. We represent them using four values: the horizontal and vertical coordinates of the top-left corner, as well as the width and height. Note that the bounding boxes provide detailed structure information like the DOM tree but in a more concise form.

\textbf{Category Features.} We store the hyperlinks of images and other types of hyperlinks. Two boolean values, \textbf{\textit{isImage}} and \textbf{\textit{isAnchor}}, indicate whether a token belongs to an image or is a hyperlink.

\textbf{Preceding Token Features.} The presence of preceding whitespace or line breaks before a text in a webpage indicates the association with its surrounding content. For instance, a line break may indicate the beginning of a new paragraph or the start of important content, and the whitespace may suggest a certain separation or hierarchical relationship. These contextual cues can aid in understanding and processing webpages. Thus, we also leverage two boolean values, \textbf{\textit{IsPrecededByLineBreak}} and \textbf{\textit{IsPrecededByWS}}, to signify whether there is a line break or whitespace before a token.

\textbf{Clickability \& Visibility Features.} Some elements in a webpage are clickable or invisible.

We use two boolean variables, \textbf{\textit{IsClipped}} and \textbf{\textit{IsVisible}}, to denote them.

All the above hypertext features form a 20-dimension array. In conjunction with them, HEED encompasses the entirety of webpage content.

\subsection{Entity Annotation}
As main product image, product name and price are the most crucial information on an e-commerce website and HEED may be intended for specific goals such as product recommendations or price monitoring, we only annotate them as shown in Figure \ref{fig:main_amazon} (a). We have hired a team of data annotators proficient in multiple languages. Specifically, they locate the entities on the webpages and annotate their corresponding spans in the extracted text, indicating the starting and ending positions. More details about \textit{HEED} are in the Appendix \ref{app:dataset_details}.

\begin{figure*}[!tp]
    \centering
    \includegraphics[width=0.95\linewidth,scale=1.00]{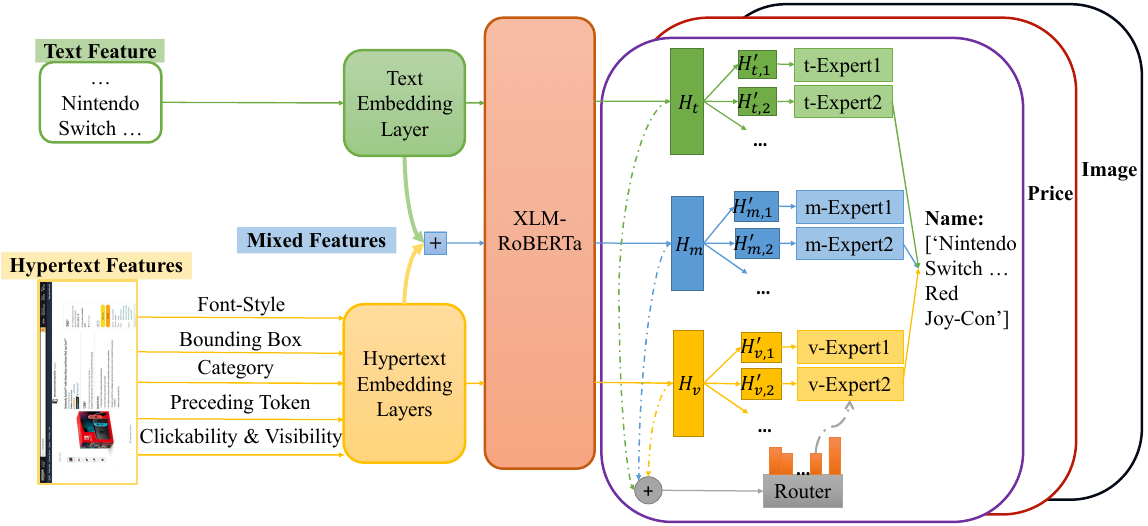}
    \caption{Overview of the MoEEF. The Hypertext Features are extracted from the original rendered webpages.}
    \label{fig:main_model}
\end{figure*}

\section{MoEEF}

MoE-based Entity Extraction Framework utilizes Mixture of Experts (MoE) to fuse multiple features as shown in Figure \ref{fig:main_model}. We can regard text and hypertext features as distinct modalities as they respectively focus on textual and visual contents. Our motivation of using MoE is that the proposed hypertext features can naturally utilize PrLM for encoding, allowing them to be combined with text features as inputs, generating various representations which naturally leads us to consider using multiple experts to make predictions based on them.

\subsection{Multi-Modal Encoding}

For each text $T=x_1,x_2,\cdots,x_n$ of length $n$ and its corresponding hypertext features $V=v_1,v_2,\cdots,v_n$ in HEED, where $v_i=v_{i1}, v_{i2},\cdots, v_{i20}$ is a vector of 20 hypertext features, MoEEF first applies text embedding layer to encode the text feature. As hypertext features have been extracted and integerized in HEED, they can be directly encoded with multiple embedding layers. We obtain the embedding of each $v_{ij}$ in $v_i$ and sum them up as the embedding of $v_i$. Subsequently, we get the text embedding $E=e_1,e_2,\cdots,e_n$ and the corresponding hypertext embedding $R=r_1,r_2,\cdots,r_n$. Furthermore, in many multi-modal information extraction models, it is common to use additional layers or modules to fuse different modalities. However, we reduce the complexity of this fusion process and straightly add the embeddings of the two modalities together to form the mixed features, enhancing the integration between modalities. Namely, MoEEF takes three modal features: \textbf{Text Feature}, \textbf{Mixed Features}, and \textbf{Hypertext Features}.

As HEED is a multi-lingual dataset, we proceed to encode the three features by leveraging powerful multi-lingual XLM-RoBERTa \cite{conneau2019unsupervised}, resulting in representations $H_t=h_{t1},h_{t2},\cdots,h_{tn}$, $H_m=h_{m1},h_{m2},\cdots,h_{mn}$ and $H_v=h_{v1},h_{v2},\cdots,h_{vn}$ for Text Feature, Mixed Features and Hypertext Features, respectively.

\subsection{MoE Decoding} \label{sec: moe_dec}
We allocate $L$ experts to each modality to independently extract entities and employ a router for a soft voting strategy to integrate their predictions. Specifically, we consider entity extraction as a sequence labeling and token-level binary classification task. For each entity (such as ``Name'' in Figure \ref{fig:main_model}), we perform binary classification on each token to determine whether it is an entity or not. To prevent the experts within the same modality from learning highly correlated knowledge, we apply multiple MLP (Multi-layer Perceptron) projectors to generate dedicated representations such as $H_{t,1}^{\prime}$ and $H_{t,2}^{\prime}$, which share the same dimension of $H_{t}$, allowing each expert to make a prediction based on its unique representation.

For the $l$-th expert in the $o \in \{t, m, v\}$ (\textit{text}, \textit{mixed}, \textit{hypertext}) modality $o\text{-Expert}l$, it is a two-layer MLP classifier with a softmax layer:
\begin{equation}\label{eq: pred}
P_{o,l} = o\text{-Expert}l(H_{o,l}^{\prime}) \in \mathbb{R}^{n \times 2}
\end{equation}

This sequence labeling approach has the advantage of handling long text on the webpage, which often needs to be divided into multiple inputs with a maximum length acceptable by the Pre-trained Language Model (PrLM). If we adopt a span extraction plan such as W$^2$NER \cite{li2022unified}, it requires concatenating all start and end positions, resulting in a quadratic storage space complexity to the input length, consuming large GPU memory during training and inference.

Then, we employ a router to score the prediction of each expert. In particular, the router is also a two-layer MLP with a softmax layer, which takes the representations of the three modalities as input and assigns a score to each expert:
\begin{equation}\label{eq: router}
\alpha = \text{Router}([H_t, H_m, H_v])
\end{equation}

The final prediction comes from the soft voting:
\begin{equation}\label{eq: fina_pred}
P_\text{final} = \sum_{o \in \{t, m, v\}}\sum_{l}^L \alpha_{o,l} P_{o,l} \in \mathbb{R}^{n \times 2}
\end{equation}

Finally, we can utilize the argmax operation to decode $P_\text{final}$ and decide whether each token has been predicted as an entity.

\begin{table*}[!tp]
\setlength\tabcolsep{1pt} 
\small
    \centering
    \begin{tabular}{c|c|ccc|ccc|ccc|ccc|ccc|ccc}
    \toprule[0.7pt]
    \midrule
    \multirow{2}{*}{Method} & \multirow{2}{*}{Task} & \multicolumn{3}{c|}{En} & \multicolumn{3}{c|}{Ar} & \multicolumn{3}{c|}{Es} & \multicolumn{3}{c|}{Ja} & \multicolumn{3}{c|}{De} & \multicolumn{3}{c}{Fr} \\
    \cline{3-20}
    
    & & P & R &  F$_1$ & P & R & F$_1$ & P & R & F$_1$ & P & R & F$_1$ & P & R & F$_1$ & P & R & F$_1$ \\
    \midrule
    \rowcolor{blue!10} \cellcolor{white} W$^2$NER(T)  &  Overall & \multicolumn{18}{c}{90.97 \quad 91.29 \quad 91.13} \\
    \midrule
    \rowcolor{blue!10} \cellcolor{white} W$^2$NER(T+V)  &  Overall & \multicolumn{18}{c}{91.76 \quad 92.53 \quad 92.14} \\

    \midrule
    \midrule
    \multirow{4}{*}{Base(T)} & Price & 91.34 & 97.07 & 94.12 & 96.77 & 93.75 & 95.24 & 93.94 & 93.94 & 93.94 & 81.48 & 84.62 & 83.02 & 85.09 & 92.38 & 88.58 & 85.71 & 84.11 & 84.91  \\
            \cmidrule{2-20}
            & Name & \textbf{95.56} & 99.16 & \textbf{97.33} & 90.91 & 93.75 & 92.31 & 100 & 100 & 100 & 100 & 100 & 100 & 88.57 & 91.18 & 89.86 & 91.07 & 95.33 & 93.15 \\
            \cmidrule{2-20}
            & Image  & 95.10 & 85.46 & 90.02 & 96.77 & 90.91 & 93.75 & 100 & 88.24 & 93.75 & \textbf{93.33} & 93.33 & 93.33 & 89.53 & 77.00 & 82.80 & 93.00 & 88.57 & 90.73 \\
            \cmidrule{2-20}
            \rowcolor{blue!10} \cellcolor{white} &  Overall & \multicolumn{18}{c}{90.78 \quad 89.87 \quad 90.33} \\

    \midrule
    \multirow{4}{*}{Base(T+V)} & Price& 82.22 & 92.89 & 87.23 & 96.77 & 93.75 & 95.24 & 88.89 & 96.97 & 92.75 & \textbf{81.48} & 84.62 & \textbf{83.02} & 78.23 & 92.38 & 84.72 & 88.18 & 90.65 & 89.40 \\
            \cmidrule{2-20}
            & Name & 87.36 & 98.33 & 92.52 & 96.88 & 96.88 & 96.88 & 91.67 & 97.06 & 94.29 & 93.55 & 96.67 & 95.08 & 82.05 & 94.12 & 87.67 & 85.71 & 95.33 & 90.27 \\
            \cmidrule{2-20}
            & Image & 85.17 & \textbf{98.68} & 91.43 & 100 & \textbf{100} & \textbf{100} & 94.29 & \textbf{97.06} & 95.65 & 90.62 & \textbf{96.67} & \textbf{93.55} & 89.36 & 84.00 & 86.60 & 82.35 & 93.33 & 87.50  \\
            \cmidrule{2-20}
             \rowcolor{blue!10} \cellcolor{white} &  Overall & \multicolumn{18}{c}{88.23 \quad \textbf{94.51} \quad 91.26} \\
    \midrule
    \midrule
    
    \multirow{4}{*}{\textbf{MoEEF}} & Price & \textbf{93.55} & \textbf{97.07} & \textbf{95.28} & \textbf{96.88} & \textbf{96.88} & \textbf{96.88} & \textbf{96.97} & \textbf{96.97} & \textbf{96.97} & 74.19 & \textbf{88.46} & 80.70 & \textbf{91.51} & \textbf{92.38} & \textbf{91.94} & \textbf{94.29} & \textbf{92.52} & \textbf{93.40} \\
            \cmidrule{2-20}
            & Name & 94.82 & \textbf{99.58} & 97.14 & \textbf{96.88} & \textbf{96.88} & \textbf{96.88} & \textbf{100} & \textbf{100} & \textbf{100} & \textbf{100} & \textbf{100} & \textbf{100} & \textbf{91.43} & \textbf{94.12} & \textbf{92.75} & \textbf{93.69} & \textbf{97.20} & \textbf{95.41} \\
            \cmidrule{2-20}
            & Image & \textbf{96.05} & 96.48 & \textbf{96.26} & \textbf{100} & 93.94 & 96.88 & \textbf{100} & 94.12 & \textbf{96.97} & 92.00 & 76.67 & 83.64 & \textbf{90.43} & \textbf{85.00} & \textbf{87.63} & \textbf{96.12} & \textbf{94.29} & \textbf{95.19}  \\
            \cmidrule{2-20}
             \rowcolor{blue!10} \cellcolor{white} &  Overall & \multicolumn{18}{c}{\textbf{94.76} \quad 94.40 \quad \textbf{94.58}} \\
    \midrule
    \bottomrule
    \end{tabular}
    \caption{Main results on multiple tasks and languages. W$^2$NER is the SOTA entity extraction model. Base refers to the vanilla XLM-RoBERTa. (T) and (T+V) represent feeding only text and the text as well as hypertext features.}
    \label{tab:main_res}
\end{table*}

\subsection{Multi-task Training}

As shown in Section \ref{sec: moe_dec}, we utilize MoE to integrate various binary classifiers for the prediction of each entity type. We treat the recognition of one type of entity as one task and there are three tasks: ``Name'', ``Price'' and ``Image''.

For a task $q \in \{\text{Name}, \text{Price}, \text{Image}\}$, we aim to train each expert to accurately predict the entities. Assuming the golden labels for one task as $G$, we use the cross-entropy loss to train each expert:

\begin{equation}
\begin{aligned}
\mathcal{L}_{q,1}=-\frac{1}{3L} \sum_{o \in \{t, m, v\}}\sum_{l}^L\sum_{i}^{n} G_{i} \log{P_{o,l,i}}
\end{aligned}
\end{equation}

To improve the accuracy of the final predictions for a task, we also utilize cross-entropy loss to refine the final prediction directly:

\begin{equation}
\begin{aligned}
\mathcal{L}_{q,2}=-\sum_{i}^{n} G_{i} \log{P_{\text{final},i}}
\end{aligned}
\end{equation}

The total loss for this task $q$ is:

\begin{equation}
\begin{aligned}
\mathcal{L}_{q}=\beta_{q,1}\mathcal{L}_{q,1} + \beta_{q,2}\mathcal{L}_{q,2}
\end{aligned}
\end{equation}

And we train the MoEEF for all the tasks in a multi-task way:

\begin{equation}
\begin{aligned}
\mathcal{L}_{total}=\sum_{q}\mathcal{L}_{q}
\end{aligned}
\end{equation}

\section{Experiments}
\subsection{Dataset and Metric}

For efficiency, we randomly select one-tenth of the HEED for experiments. We split 80\% data as the training set, 10\% as the development set, and 10\% for test. We use precision (P), recall (R), and F$_1$ as metrics following common practices.

\subsection{Settings}
We leverage the XLM-RoBERTa-base \cite{conneau2019unsupervised} as the backbone. Due to the typically long length of web texts, we divide them into different sequences with a maximum length of 512. We configure each modality to have 6 experts. For training, we use the AdamW \cite{loshchilov2017decoupled} optimizer with a learning rate of 1e-5. The framework is trained on 8 NVIDIA V100 GPUs for 5 epochs on the training set with a batch size of 96. We select the checkpoint with the highest F$_1$ score on the development set for the test. Empirically, we assign the two coefficients for loss $\mathcal{L}_{q}$ as $\beta_{q,1}=0.8$ and $\beta_{q,2}=0.2$.

\subsection{Main Results}

Our main results are shown in Table \ref{tab:main_res}. We reproduce the W$^2$NER \cite{li2022unified} on the HEED, which is the state-of-the-art (SOTA) entity extraction model for comparison\footnote{Our dataset is not suitable for Webformer \cite{wang2022webformer} and FreeDOM \cite{lin2020freedom} as they require the DOM tree.}. We train it in a multi-task fashion with three classification heads to individually recognize the entities of Price, Name, and Image. To ensure fairness, its backbone model is XLM-RoBERTa-base. We present its overall P, R, and F$_1$ for the three tasks. The (T) and (T+V) respectively represent feeding only text features and the sum of text and hypertext features.

We employ XLM-RoBERTa-base in combination with text or text and hypertext features to extract different entities in a multi-task fashion, shown as Base(T) and Base(T+V). We present the results on several popular languages with a large number of samples and domains.

Overall, our MoEEF significantly outperforms other baselines in terms of F$_1$. Even though slightly lower recall than Base(T+V), MoEEF exhibits more advantageous precision.

Comparing W$^2$NER(T) with W$^2$NER(T+V) and Base(T) with Base(T+V), we can see an obvious increase in F$_1$, which demonstrates the effectiveness of the extracted hypertext features in HEED.

In addition, from the perspective of different languages, MoEEF also demonstrates clear advantages, as its consistently higher P, R, and F$_1$. Furthermore, we find that the MoEEF can decrease the distinction between P and R for most languages. We calculate the average of the absolute differences between each P and its corresponding R in Table \ref{tab:main_res}, as shown in Table \ref{tab:abs_mean}. MoEEF generally achieves a smaller difference, which shows that it better balances P and R for improvement on F$_1$.

Nowadays, the large language models (LLMs) have shown impressive performance in diverse tasks. We conduct experiments with one of the SOTA GPT-3.5-turbo model on the English test set of \textit{HEED} dataset with the zero-shot prompts detailed in Appendix \ref{app:prompt}. Results in Table \ref{tab:gpt3.5} indicate a notable performance gap compared to Table \ref{tab:main_res}. This aligns with prior challenges faced by LLMs in surpassing BERT-like models in NER tasks \cite{wang2023gpt}. Fortunately, Table \ref{tab:gpt3.5} highlights performance improvements with the inclusion of the hypertext feature, emphasizing its effectiveness.

\subsection{Analysis of Router}

\begin{figure}[!tp]
    \centering
    \includegraphics[width=0.95\linewidth,scale=1.00]{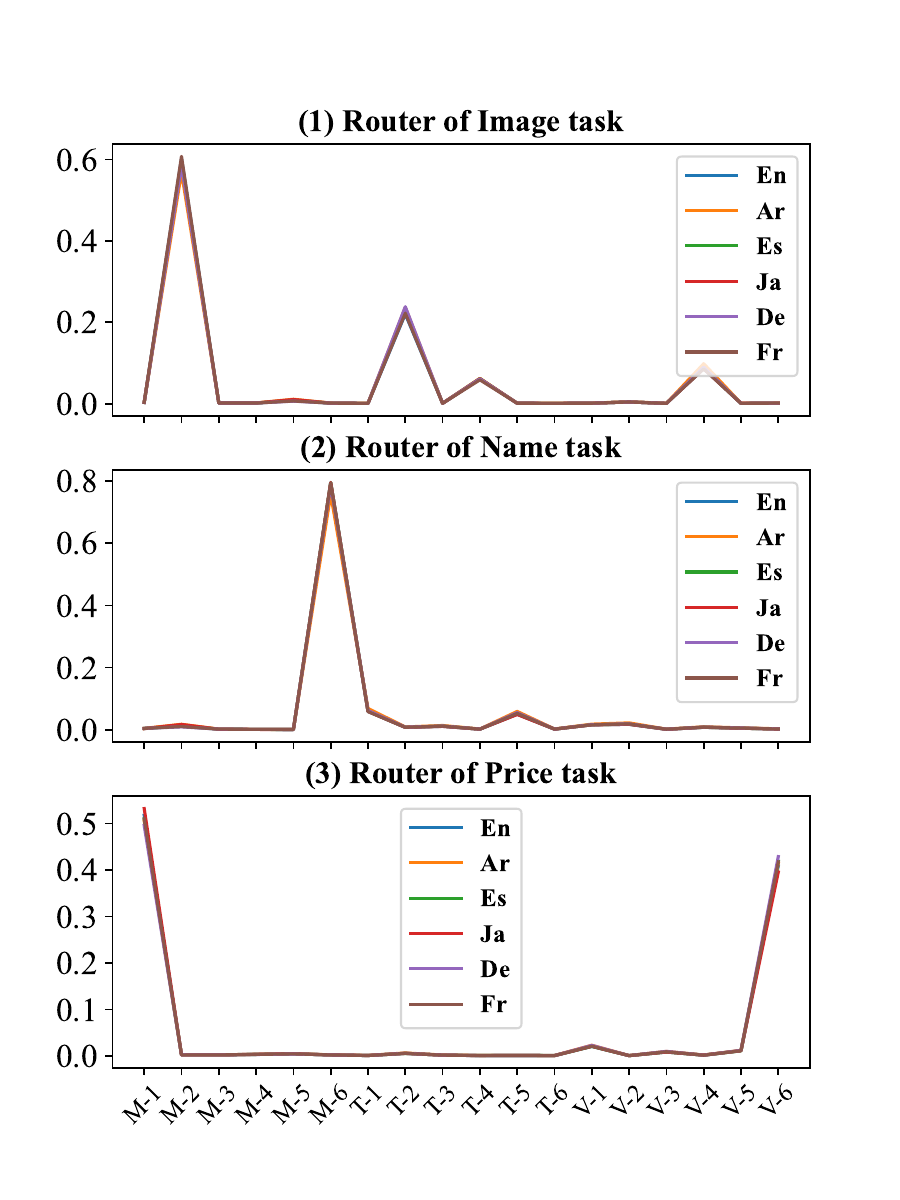}
    \caption{Visualization of the router for different tasks.}
    \label{fig:router_vis}
\end{figure}

The router is crucial by facilitating the soft integration of final predictions. We can understand the contributions of different experts and features by analyzing it. For one specific task, we feed all samples from one language into MoEEF to obtain the probability distributions of the router on all experts and then calculate their average values. We plot the probability distributions of different languages for the same task on one line graph, as shown in Figure \ref{fig:router_vis}, where ``M'', ``T'', and ``V'' correspond to the experts for mixed, text, and hypertext experts.

Surprisingly, for each task, the router displays remarkable similarity in selecting experts for samples from different languages which indicates that the output of the router is highly task-specific and does not rely on the languages. For Image extraction, all of the mixed, hypertext, and text features are very useful. For the Name, mixed features play a major role, while both mixed and hypertext features are helpful for Price entity extraction. Meanwhile, it also reflects that multi-lingual language models such as XLM-RoBERTa can encode texts from different languages into the same semantic space.

\begin{table}[!tp]
\setlength\tabcolsep{12pt}
    \tiny
    \centering
    \scriptsize
    \resizebox{0.95\linewidth}{!}{
    \begin{tabular}{lc}
         \toprule
         \midrule
         \bf Method & Absolute Difference (Avg.)\\
         \midrule
         Base(T) & 4.35 \\
        Base(T+v) & 6.74  \\
        \bf MoEEF & \textbf{3.69}  \\
         \midrule
         \bottomrule
    \end{tabular}}
    \caption{The average absolute differences between each P and R.}
    \label{tab:abs_mean}
\end{table}

\begin{table}[!tp]
\setlength\tabcolsep{7pt}
    \tiny
    \centering
    \scriptsize
    \resizebox{0.95\linewidth}{!}{
    \begin{tabular}{lcccc}
         \toprule
         \midrule
         \bf Model & & P & R & F$_1$ \\
         \midrule
         \multirow{3}{*}{GPT3.5(T)} & Price & 2.11 & 0.74 & 1.10 \\
         & Name & 7.49 & 3.36 & 4.64 \\
         & Image & 17.38 & 4.58 & 7.25 \\
         \midrule
         \multirow{3}{*}{GPT3.5(T+V)} & Price & 48.33 & 5.87 & 10.45 \\
         & Name & 70.18 & 22.60 & 34.19 \\
         & Image & 61.49 & 10.00 & 17.20 \\
         \midrule
         \bottomrule
    \end{tabular}}
    \caption{The results of GPT-3.5-turbo on the English test set of HEED.}
    \label{tab:gpt3.5}
\end{table}

\begin{figure*}[!tp]
    \centering
    \includegraphics[width=\linewidth,scale=0.9]{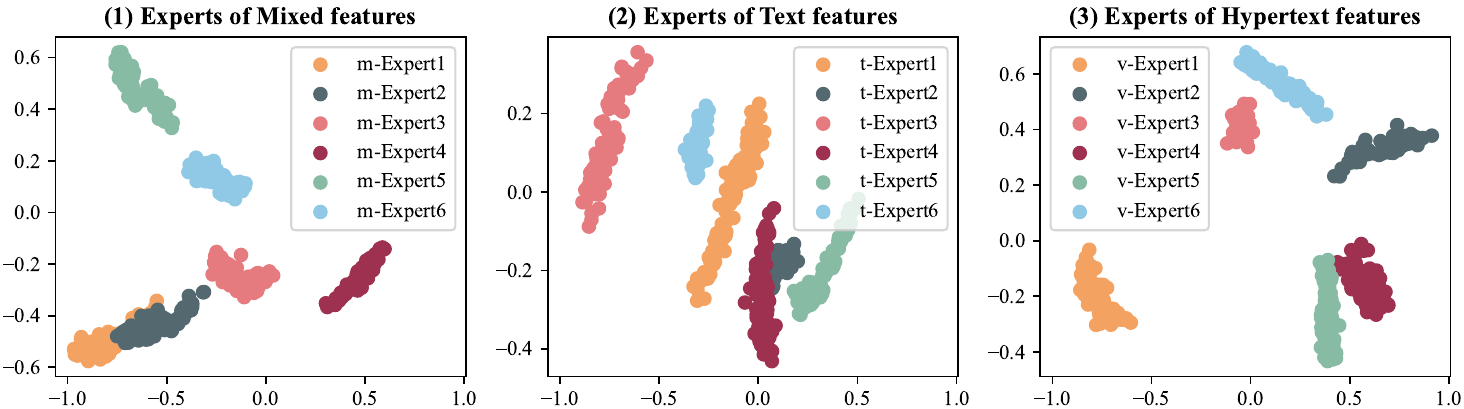}
    \caption{Visualizations of representations for different experts.}
    \label{fig:repr_vis}
\end{figure*}

\subsection{Distinctiveness of Expert Representations}

\begin{table}[!tbp]
\setlength\tabcolsep{10pt}
    \centering
    \scriptsize
    \resizebox{0.95\linewidth}{!}{
    \begin{tabular}{lccc}
         \toprule
         \midrule
         \bf Method & P & R & F$_1$ \\
         \midrule
         \bf MoEEF & \textbf{94.76} & \textbf{94.40} & \textbf{94.58} \\
         \quad+$L_{ort}$ & 94.40 & 93.24 & 93.81(-0.77) \\
         \midrule
         \bottomrule
    \end{tabular}}
    \caption{Result with Orthogonal Regularization Loss.}
    \label{tab:diff_exp}
\end{table}

The representations $H_{o,1}^{\prime},H_{o,2}^{\prime},\cdots,H_{o,L}^{\prime}$ of one modality $o$ are the dependence for corresponding experts to make their predictions. However, we do not prevent excessive similarity among the them, which could result in too consistent experts and consequently rendering the MoE ineffective.

Inspired by \citet{ranasinghe2021orthogonal}, we attempt to introduce differentiation constraint among these representations by \textit{Orthogonal Regularization Loss}. For the $H_{o,l}^{\prime} \in \mathbb{R}^{n \times d}$ of the modality $o$, we stack them into one matrix and normalize it to get $\mathcal{H}_o \in \mathbb{R}^{n \times L \times d}$. Then we transpose its last two dimensions to get $\mathcal{H}_o^{\top} \in \mathbb{R}^{n \times d \times L}$. The Orthogonal Regularization Loss for this modality is:
\begin{equation}
L_{ort}(\mathcal{H}_o)=\| \operatorname{matmul}\left(\mathcal{H}_o, \mathcal{H}_o^{\top}\right)-\mathcal{I} \|_2
\end{equation}
where $\mathcal{I} \in \mathbb{R}^{n \times L \times L}$ is the identity matrix. 

A smaller $L_{ort}$ can make each $H_{o,l}^{\prime}$ more orthogonal to the others, thereby reducing their similarity and differentiating the experts. We calculate a separate loss for each modality and then optimize them together with the MoEEF loss function. The result is shown in Table \ref{tab:diff_exp}. Unfortunately, we observe a notable decrease in P, R, and F$_1$.

Such a phenomenon hints us to investigate the original $H_{o,l}^{\prime}$ when the $L_{ort}$ is not applied. We randomly select 16 samples for Price in different languages and input them into well-trained MoEEF to obtain the input representations $H_{o,l}^{\prime}$ of the three modalities. We visualize them after dimension-reduction by t-SNE \cite{van2008visualizing}, as in Figure \ref{fig:repr_vis}\footnote{The visualization of all three tasks are in the Appendix \ref{app:vis}}. The input representations of different experts in each modality form distinct clusters, indicating that they have spontaneously learned different knowledge and behave variously, which satisfies our demand for expert differentiation. Thus, imposing additional differentiation constraints may be not necessary.

\section{Ablation Study}

\subsection{Effect of Hypertext Features}

As mentioned in Sec.\ref{sec:vis feat}, there are 5 categories hypertext features: \textit{Font-Style}, \textit{Bounding Box}, \textit{Category}, \textit{Preceding Token} and \textit{Clickability \& Visibility}. We intend to investigate their impact on MoEEF. Specifically, we discard each of them and re-train the MoEEF, as shown in Table \ref{tab:abl_visfeat}.

The F$_1$ decreases after removing any of the hypertext features, indicating that each one contributes to the performance. Moreover, the \textit{Font-Style} and \textit{Bounding Box} features are most helpful as expected. Clearly, they provide rich information such as font, color, and position, which are also highly informative to humans. The experiments also demonstrate that the \textit{Category} and \textit{Preceding Token} are informative, as they contribute to an increase of 1-2 F$_1$. The \textit{Clickability \& Visibility} show some performance improvement as well.

\begin{table}[!tbp]
\setlength\tabcolsep{7pt}
    \centering
    \scriptsize
    \resizebox{0.95\linewidth}{!}{
    \begin{tabular}{lccc}
         \toprule
         \midrule
         \bf Modality & P & R & F$_1$ \\
         \midrule
         \bf All & \textbf{94.76} & \textbf{94.40} & \textbf{94.58} \\
         \midrule
         \quad-Hypertext & 93.68 & 94.71 & 94.19(-0.39) \\
         \quad-Text & 90.46 & 94.02 & 92.21(-2.37) \\
         \quad-Mixed & 93.34 & 89.74 & 91.50(-3.08)\\
         \midrule
         \bottomrule
    \end{tabular}}
    \caption{Result without each modal input.}
    \label{tab:modal_input}
\end{table}

\begin{table}[!tbp]
\setlength\tabcolsep{4pt}
    \centering
    \small
    \resizebox{0.95\linewidth}{!}{
    \begin{tabular}{lccc}
         \toprule
         \midrule
         \bf Vis Feats. & P & R & F$_1$ \\
         \midrule
         \bf All & \textbf{94.76} & \textbf{94.40} & \textbf{94.58} \\
         \midrule
         \quad-Font Style & 93.40 & 89.74 & 91.50(-3.08) \\
         \quad-Bounding Box & 93.49 & 88.69 & 91.03(-3.55) \\
         \quad-Category & 94.17 & 92.26 & 93.20(-1.38) \\
         \quad-Preceding Token & 93.76 & 91.95 & 92.85(-1.73) \\
         \quad-Clickability \& Visibility & 95.20 & 93.24 & 94.21(-0.37) \\
         \midrule
         \bottomrule
    \end{tabular}}
    \caption{Ablation study on hypertext features.}
    \label{tab:abl_visfeat}
\end{table}

\subsection{Effect of Multi-Modal Input}

To verify that multi-modal input can benefit MoEEF and determine which modality is most practical, we keep settings unchanged and re-train the MoEEF while removing each modal input in turn. The results are shown in Table \ref{tab:modal_input}.

It can be observed that after removing the inputs of the three modalities separately, all P, R, and F$_1$ decrease significantly, which indicates that each modality is helpful. Moreover, We find that hypertext features and text feature have a smaller impact compared to mixed features. This suggests that the additional mixed features we introduced are highly valuable. This is also evident from Figure \ref{fig:router_vis}, where each task heavily relies on mixed features for entity extraction. We further speculate that fusion features are more precious for multi-modal models.

\subsection{Effect of Expert Amounts}

The amount of experts for each modality is an important hyper-parameter for MoEEF, and we conduct ablation studies on it. We sequentially set the number of experts for each modality to 1, 3, 6, and 9 and re-train the framework numerous times. The results are shown in Table \ref{tab:expert_amount}. Increasing the number of experts from 1 to 6 results in improved performance. However, when it exceeds 6, the performance starts to decline. This may suggest that more experts are not necessarily better for the MoE system. Having too many experts introduces excessive parameters while also negatively impacting performance. The parameter for the experts amount may be an empirical setting.

\section{Related Work}
Webpage entity extraction has been extensively studied in academia and practical applications. The common practice is to train specialized models \cite{wang2022webformer,nasar2021named,cao2021extracting}. Although LLMs demonstrate impressive performance across various tasks, their performance in information extraction (IE) is superior \cite{zhong2023can, zhao2023survey}, which is also verified by our experiment. 
As their deployment cost also remains unacceptably high, specialized small-scale models still have value today.

\begin{table}[!tbp]
\setlength\tabcolsep{9pt}
    \centering
    \scriptsize
    \resizebox{0.95\linewidth}{!}{
    \begin{tabular}{cccc}
         \toprule
         \midrule
         \bf Experts & P & R & F$_1$ \\
         \midrule
         \bf 6 & \textbf{94.76} & \textbf{94.40} & \textbf{94.58} \\
         \midrule
         1 & 94.15 & 94.17 & 94.16(-0.42) \\
         3 & 93.73 & 94.82 & 94.27(-0.31) \\
         9 & 94.37 & 93.89 & 94.13(-0.45)\\
         \midrule
         \bottomrule
    \end{tabular}}
    \caption{Ablation study on experts amount.}
    \label{tab:expert_amount}
\end{table}

As for datasets, SWDE \cite{hao2011one} is one of the earliest datasets, collecting over 100,000 webpages from 8 domains. The Expanded SWDE \cite{lockard2019openceres} expands on it by adding 3 domains. WEIR \cite{bronzi2013extraction} is another early structured webpage IE dataset focusing on overlapping data from different sources. Common Crawl\footnote{http://commoncrawl.org/connect/blog/} is a widely used large-scale dataset for IE. However, they all overlook rich hypertext features and contain too much noise. Moreover, their annotations are often determined by rules or automated tools, lacking accurate manual annotations.

Based on the above datasets, \citet{lockard2020zeroshotceres} propose a zero-shot webpage IE model with strong generalization. LayoutLM \cite{xu2020layoutlm} utilizes layout and style information to pre-train a model. \citet{lin2020freedom} mine webpage information through the representation of DOM tree nodes by a relational neural network. \citet{zhou2021simplified} efficiently retrieve useful context for each DOM node. \citet{wang2022webformer} also need the representation of the DOM node for IE. While they have made some progress, they all require additional components such as GNN \cite{scarselli2008graph}, GAT \cite{velivckovic2017graph}, Faster R-CNN \cite{girshick2015fast}, or separate modeling of the text DOM tree, leading to overly complex models.

\section{Conclusion}
Existing webpage entity recognition datasets only retain the text and structure information, overlooking the rich hypertext features. This paper first introduces a dataset called \textit{HEED} that explicitly extracts rich hypertext features. Moreover, we develop a entity extraction framework called \textit{MoEEF}, based on the Mixture of Experts, which significantly outperforms strong baselines, including the state-of-the-art small-scale models and GPT-3.5-turbo. Detailed ablation studies and analysis prove the effectiveness of extracted hypertext features in \textit{HEED} and several model components.

\section*{Limitations}
We do not extensively explore why the input representations of different experts would exhibit significant variations spontaneously in an unconstrained setting and leave it as future work. Moreover, this paper does not consider the issue of balancing the MoE experts. Our router demonstrates that only a few of experts are contributing while balancing the experts is effective in some MoE systems \cite{shazeer2017outrageously}. We regard how to balance the experts in the MoEEF as another future work. The currently advanced multimodal GPT-4V model is still under restricted access, and we are temporarily unable to evaluate it. We will also include more experiments with advanced large language models in future updates.

\bibliography{acl_latex}

\begin{thebibliography}{25}
\expandafter\ifx\csname natexlab\endcsname\relax\def\natexlab#1{#1}\fi

\bibitem[{Bhardwaj et~al.(2021)Bhardwaj, Ahmed, Jaiharie, Dadhich, and Ganesan}]{bhardwaj2021web}
Bhavya Bhardwaj, Syed~Ishtiyaq Ahmed, J~Jaiharie, R~Sorabh Dadhich, and M~Ganesan. 2021.
\newblock Web scraping using summarization and named entity recognition (ner).
\newblock In \emph{2021 7th international conference on advanced computing and communication systems (ICACCS)}, volume~1, pages 261--265. IEEE.

\bibitem[{Bronzi et~al.(2013)Bronzi, Crescenzi, Merialdo, and Papotti}]{bronzi2013extraction}
Mirko Bronzi, Valter Crescenzi, Paolo Merialdo, and Paolo Papotti. 2013.
\newblock Extraction and integration of partially overlapping web sources.
\newblock \emph{Proceedings of the VLDB Endowment}, 6(10):805--816.

\bibitem[{Cao and Luo(2021)}]{cao2021extracting}
Rongyu Cao and Ping Luo. 2021.
\newblock Extracting zero-shot structured information from form-like documents: Pretraining with keys and triggers.
\newblock In \emph{Proceedings of the AAAI Conference on Artificial Intelligence}, volume~35, pages 12612--12620.

\bibitem[{Chen et~al.(2009)Chen, Lee, and Huang}]{chen2009polyuhk}
Ying Chen, S~Yat~Mei Lee, and Chu-Ren Huang. 2009.
\newblock Polyuhk: A robust information extraction system for web personal names.
\newblock In \emph{2nd Web People Search Evaluation Workshop (WePS 2009), 18th WWW Conference}.

\bibitem[{Conneau et~al.(2020)Conneau, Khandelwal, Goyal, Chaudhary, Wenzek, Guzm{\'a}n, Grave, Ott, Zettlemoyer, and Stoyanov}]{conneau2019unsupervised}
Alexis Conneau, Kartikay Khandelwal, Naman Goyal, Vishrav Chaudhary, Guillaume Wenzek, Francisco Guzm{\'a}n, Edouard Grave, Myle Ott, Luke Zettlemoyer, and Veselin Stoyanov. 2020.
\newblock \href {https://doi.org/10.18653/v1/2020.acl-main.747} {Unsupervised cross-lingual representation learning at scale}.
\newblock In \emph{Proceedings of the 58th Annual Meeting of the Association for Computational Linguistics}, pages 8440--8451, Online. Association for Computational Linguistics.

\bibitem[{Girshick(2015)}]{girshick2015fast}
Ross~B. Girshick. 2015.
\newblock \href {https://doi.org/10.1109/ICCV.2015.169} {Fast {R-CNN}}.
\newblock In \emph{2015 {IEEE} International Conference on Computer Vision, {ICCV} 2015, Santiago, Chile, December 7-13, 2015}, pages 1440--1448. {IEEE} Computer Society.

\bibitem[{Hao et~al.(2011)Hao, Cai, Pang, and Zhang}]{hao2011one}
Qiang Hao, Rui Cai, Yanwei Pang, and Lei Zhang. 2011.
\newblock \href {https://doi.org/10.1145/2009916.2010020} {From one tree to a forest: a unified solution for structured web data extraction}.
\newblock In \emph{Proceeding of the 34th International {ACM} {SIGIR} Conference on Research and Development in Information Retrieval, {SIGIR} 2011, Beijing, China, July 25-29, 2011}, pages 775--784. {ACM}.

\bibitem[{Li et~al.(2022)Li, Fei, Liu, Wu, Zhang, Teng, Ji, and Li}]{li2022unified}
Jingye Li, Hao Fei, Jiang Liu, Shengqiong Wu, Meishan Zhang, Chong Teng, Donghong Ji, and Fei Li. 2022.
\newblock Unified named entity recognition as word-word relation classification.
\newblock In \emph{Proceedings of the AAAI Conference on Artificial Intelligence}, volume~36, pages 10965--10973.

\bibitem[{Lin et~al.(2020)Lin, Sheng, Vo, and Tata}]{lin2020freedom}
Bill~Yuchen Lin, Ying Sheng, Nguyen Vo, and Sandeep Tata. 2020.
\newblock \href {https://dl.acm.org/doi/10.1145/3394486.3403153} {Freedom: {A} transferable neural architecture for structured information extraction on web documents}.
\newblock In \emph{{KDD} '20: The 26th {ACM} {SIGKDD} Conference on Knowledge Discovery and Data Mining, Virtual Event, CA, USA, August 23-27, 2020}, pages 1092--1102. {ACM}.

\bibitem[{Lockard et~al.(2019)Lockard, Shiralkar, and Dong}]{lockard2019openceres}
Colin Lockard, Prashant Shiralkar, and Xin~Luna Dong. 2019.
\newblock \href {https://doi.org/10.18653/v1/N19-1309} {{O}pen{C}eres: {W}hen open information extraction meets the semi-structured web}.
\newblock In \emph{Proceedings of the 2019 Conference of the North {A}merican Chapter of the Association for Computational Linguistics: Human Language Technologies, Volume 1 (Long and Short Papers)}, pages 3047--3056, Minneapolis, Minnesota. Association for Computational Linguistics.

\bibitem[{Lockard et~al.(2020)Lockard, Shiralkar, Dong, and Hajishirzi}]{lockard2020zeroshotceres}
Colin Lockard, Prashant Shiralkar, Xin~Luna Dong, and Hannaneh Hajishirzi. 2020.
\newblock \href {https://doi.org/10.18653/v1/2020.acl-main.721} {{Z}ero{S}hot{C}eres: Zero-shot relation extraction from semi-structured webpages}.
\newblock In \emph{Proceedings of the 58th Annual Meeting of the Association for Computational Linguistics}, pages 8105--8117, Online. Association for Computational Linguistics.

\bibitem[{Loshchilov and Hutter(2019)}]{loshchilov2017decoupled}
Ilya Loshchilov and Frank Hutter. 2019.
\newblock \href {https://openreview.net/forum?id=Bkg6RiCqY7} {Decoupled weight decay regularization}.
\newblock In \emph{7th International Conference on Learning Representations, {ICLR} 2019, New Orleans, LA, USA, May 6-9, 2019}. OpenReview.net.

\bibitem[{Nasar et~al.(2021)Nasar, Jaffry, and Malik}]{nasar2021named}
Zara Nasar, Syed~Waqar Jaffry, and Muhammad~Kamran Malik. 2021.
\newblock Named entity recognition and relation extraction: State-of-the-art.
\newblock \emph{ACM Computing Surveys (CSUR)}, 54(1):1--39.

\bibitem[{Ranasinghe et~al.(2021)Ranasinghe, Naseer, Hayat, Khan, and Khan}]{ranasinghe2021orthogonal}
Kanchana Ranasinghe, Muzammal Naseer, Munawar Hayat, Salman Khan, and Fahad~Shahbaz Khan. 2021.
\newblock Orthogonal projection loss.
\newblock In \emph{Proceedings of the IEEE/CVF International Conference on Computer Vision}, pages 12333--12343.

\bibitem[{Scarselli et~al.(2008)Scarselli, Gori, Tsoi, Hagenbuchner, and Monfardini}]{scarselli2008graph}
Franco Scarselli, Marco Gori, Ah~Chung Tsoi, Markus Hagenbuchner, and Gabriele Monfardini. 2008.
\newblock The graph neural network model.
\newblock \emph{IEEE transactions on neural networks}, 20(1):61--80.

\bibitem[{Shazeer et~al.(2017)Shazeer, Mirhoseini, Maziarz, Davis, Le, Hinton, and Dean}]{shazeer2017outrageously}
Noam Shazeer, Azalia Mirhoseini, Krzysztof Maziarz, Andy Davis, Quoc~V. Le, Geoffrey~E. Hinton, and Jeff Dean. 2017.
\newblock \href {https://openreview.net/forum?id=B1ckMDqlg} {Outrageously large neural networks: The sparsely-gated mixture-of-experts layer}.
\newblock In \emph{5th International Conference on Learning Representations, {ICLR} 2017, Toulon, France, April 24-26, 2017, Conference Track Proceedings}. OpenReview.net.

\bibitem[{Van~der Maaten and Hinton(2008)}]{van2008visualizing}
Laurens Van~der Maaten and Geoffrey Hinton. 2008.
\newblock Visualizing data using t-sne.
\newblock \emph{Journal of machine learning research}, 9(11).

\bibitem[{Velickovic et~al.(2018)Velickovic, Cucurull, Casanova, Romero, Li{\`{o}}, and Bengio}]{velivckovic2017graph}
Petar Velickovic, Guillem Cucurull, Arantxa Casanova, Adriana Romero, Pietro Li{\`{o}}, and Yoshua Bengio. 2018.
\newblock \href {https://openreview.net/forum?id=rJXMpikCZ} {Graph attention networks}.
\newblock In \emph{6th International Conference on Learning Representations, {ICLR} 2018, Vancouver, BC, Canada, April 30 - May 3, 2018, Conference Track Proceedings}. OpenReview.net.

\bibitem[{Wang et~al.(2022)Wang, Fang, Ravula, Feng, Quan, and Liu}]{wang2022webformer}
Qifan Wang, Yi~Fang, Anirudh Ravula, Fuli Feng, Xiaojun Quan, and Dongfang Liu. 2022.
\newblock Webformer: The web-page transformer for structure information extraction.
\newblock In \emph{Proceedings of the ACM Web Conference 2022}, pages 3124--3133.

\bibitem[{Wang et~al.(2023)Wang, Sun, Li, Ouyang, Wu, Zhang, Li, and Wang}]{wang2023gpt}
Shuhe Wang, Xiaofei Sun, Xiaoya Li, Rongbin Ouyang, Fei Wu, Tianwei Zhang, Jiwei Li, and Guoyin Wang. 2023.
\newblock Gpt-ner: Named entity recognition via large language models.
\newblock \emph{arXiv preprint arXiv:2304.10428}.

\bibitem[{Wong and Lam(2009)}]{wong2009learning}
Tak-Lam Wong and Wai Lam. 2009.
\newblock Learning to adapt web information extraction knowledge and discovering new attributes via a bayesian approach.
\newblock \emph{IEEE Transactions on Knowledge and Data Engineering}, 22(4):523--536.

\bibitem[{Xu et~al.(2020)Xu, Li, Cui, Huang, Wei, and Zhou}]{xu2020layoutlm}
Yiheng Xu, Minghao Li, Lei Cui, Shaohan Huang, Furu Wei, and Ming Zhou. 2020.
\newblock \href {https://dl.acm.org/doi/10.1145/3394486.3403172} {Layoutlm: Pre-training of text and layout for document image understanding}.
\newblock In \emph{{KDD} '20: The 26th {ACM} {SIGKDD} Conference on Knowledge Discovery and Data Mining, Virtual Event, CA, USA, August 23-27, 2020}, pages 1192--1200. {ACM}.

\bibitem[{Zhao et~al.(2023)Zhao, Zhou, Li, Tang, Wang, Hou, Min, Zhang, Zhang, Dong et~al.}]{zhao2023survey}
Wayne~Xin Zhao, Kun Zhou, Junyi Li, Tianyi Tang, Xiaolei Wang, Yupeng Hou, Yingqian Min, Beichen Zhang, Junjie Zhang, Zican Dong, et~al. 2023.
\newblock \href {https://arxiv.org/abs/2303.18223} {A survey of large language models}.
\newblock \emph{ArXiv preprint}, abs/2303.18223.

\bibitem[{Zhong et~al.(2023)Zhong, Ding, Liu, Du, and Tao}]{zhong2023can}
Qihuang Zhong, Liang Ding, Juhua Liu, Bo~Du, and Dacheng Tao. 2023.
\newblock \href {https://arxiv.org/abs/2302.10198} {Can chatgpt understand too? a comparative study on chatgpt and fine-tuned bert}.
\newblock \emph{ArXiv preprint}, abs/2302.10198.

\bibitem[{Zhou et~al.(2021)Zhou, Sheng, Vo, Edmonds, and Tata}]{zhou2021simplified}
Yichao Zhou, Ying Sheng, Nguyen Vo, Nick Edmonds, and Sandeep Tata. 2021.
\newblock \href {https://arxiv.org/abs/2101.02415} {Simplified dom trees for transferable attribute extraction from the web}.
\newblock \emph{ArXiv preprint}, abs/2101.02415.

\end{thebibliography}

\appendix


\section{Dataset Details} \label{app:dataset_details}
\subsection{Statistic from each language}

\begin{table}[!ht]
\centering
\tabcolsep=0.15cm
{\resizebox{0.99\linewidth}{!}{
\begin{tabular}{l|ccccccccc}
\toprule
\hline

\textbf{Language} & \textbf{En} & \textbf{Fr} & \textbf{De} & \textbf{Zh} & \textbf{It} & \textbf{Ko} & \textbf{Ja} & \textbf{Es} & \textbf{Ar} \\ \midrule
\begin{tabular}[c]{@{}l@{}}\textbf{Size (\textit{k})} \end{tabular}  & 180 & 90 & 80  & 20 & 6 & 19 & 38 & 5.9 & 24  \\ \midrule
\begin{tabular}[c]{@{}l@{}}\textbf{Domains} \end{tabular}  & 172 & 87 & 65 & 20 & 51 & 42 & 37 & 21 & 24 \\
\hline
\bottomrule
\end{tabular}}}
\caption{Statistics of the dataset from each language.}
\label{table:static}
\end{table}

\subsection{Quality Control}
We have implemented a variety of methods and strategies to guarantee the accuracy and the quality of our \textit{HEED} dataset.

For webpage data parsing and feature extraction, we develop an automated parsing tool that we have verified to be error-free.

For entity annotation, we hire several professional annotators proficient in multiple languages and require them to qualify through qualification tasks and spam detection. Prior to annotation, we set up 10 qualification tasks, and each annotator needs to pass at least 7 tasks to be eligible for entity annotation. We establish a spam set consisting of 500 samples, containing a mix of correct and spam samples. Annotators are required to correctly identify these samples as either legitimate or spam with an accuracy of at least 70\%. In practice, their accuracy on the spam set reaches 80.1\%.

For annotation consistency, each sample is annotated by 5 annotators. If more than 3 annotators provide consistent labels for a sample, that label is considered the final label. And the rate of agreement with most common is 86.7\%.

\subsection{Length Distribution}

\begin{figure}[!h]
    \centering
    \includegraphics[width=0.95\linewidth,scale=1.00]{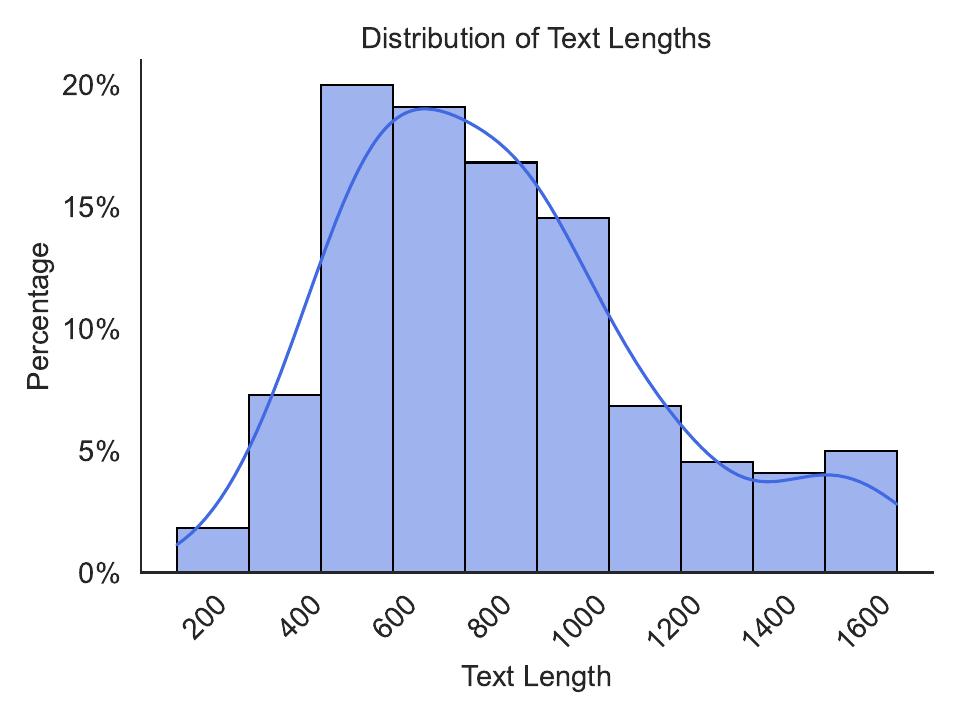}
    \caption{The distribution of the text length on the webpages.}
    \label{fig:len_dis}
\end{figure}

We analyze the distribution of text lengths in the dataset and find that the majority of sentences fall within the range of 400 to 1000 tokens, as shown in Figure \ref{fig:len_dis}. The average length is about 750 tokens.

\subsection{Entity Distribution in Sentences}
\begin{figure}[!h]
    \centering
    \includegraphics[width=0.95\linewidth,scale=1.00]{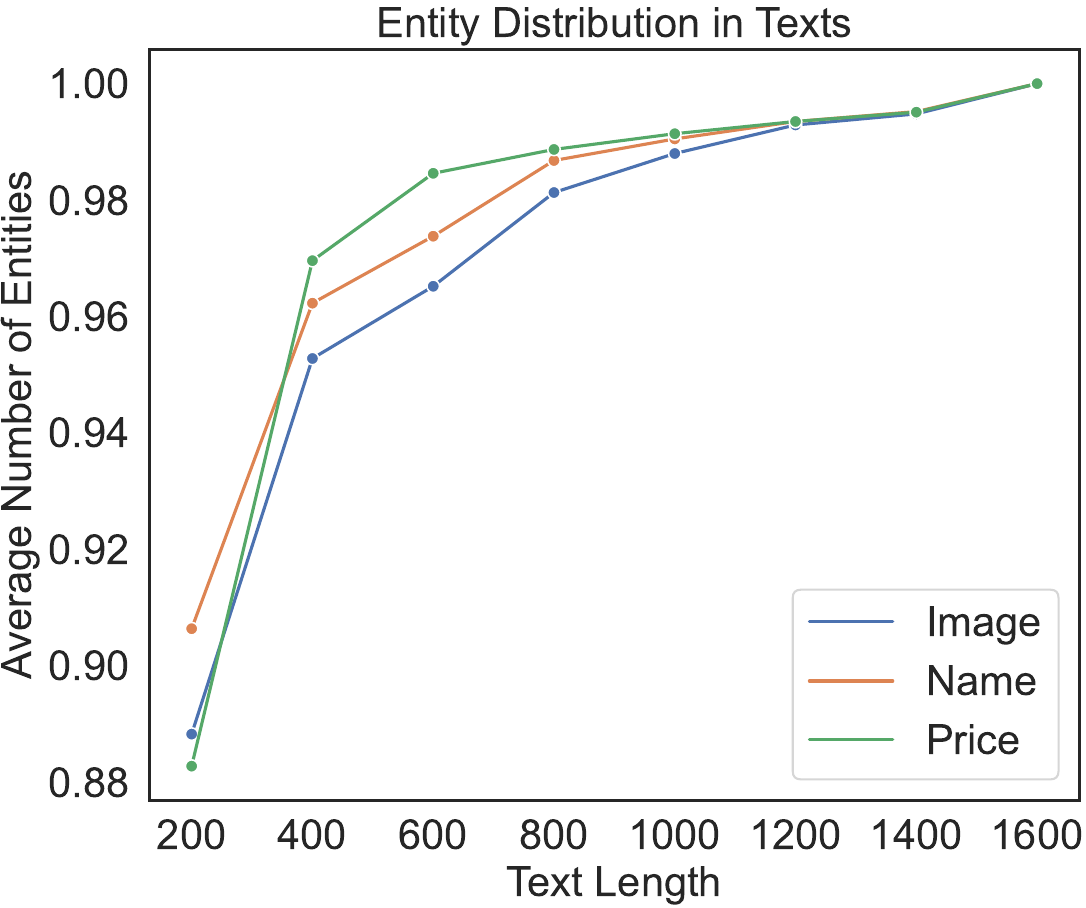}
    \caption{The entity distribution in webpage texts.}
    \label{fig:ett_dis}
\end{figure}

We also examine the distribution of entities within sentences as shown in Figure \ref{fig:ett_dis} and find that, on average, at least 88\% of entities can be located within the first 200 tokens of a webpage text. The first 800 tokens encompass 98\% of entities.

In subsequent experiments, for the texts exceeding the maximum acceptable length of the model, we segment such texts into multiple sections.

\section{Prompt For GPT-3.5-turbo} \label{app:prompt}

\begin{tcolorbox}
    \textbf{Without Hypertext Feature}\\
     Given the text of a web page: \{\}, please extract all entities of type \{\}. You need to only return the corresponding start and end positions of the spans like [(1,2), (5,6)]]. Please remember the output format and never give me any redundant information.\\
    \textbf{With Hypertext Feature}\\
    Given the text of a web page: \{\}, where each token corresponds to HTML features \{\}. Each vector corresponds to a set of hypertext features, including font-size, bounding box, and other details. Please extract all entities of type \{\}, and return the corresponding start and end positions of the spans like [(1,2), (5,6)]]. Please remember the output format and never give me any redundant information.
\end{tcolorbox}

To maximize input length within the constraints of GPT-3.5-turbo, when not using hypertext features, we split the input text every 1024 tokens. When utilizing hypertext features, the input text is split every 128 tokens, along with the corresponding hypertext feature.

We have actually tested numerous prompts, some of which may sound more natural, but the effectiveness is not satisfactory as the above.


\section{Visualization of Experts Input} \label{app:vis}
\begin{figure*}[!htp]
    \centering
    \includegraphics[width=\linewidth,scale=1.00]{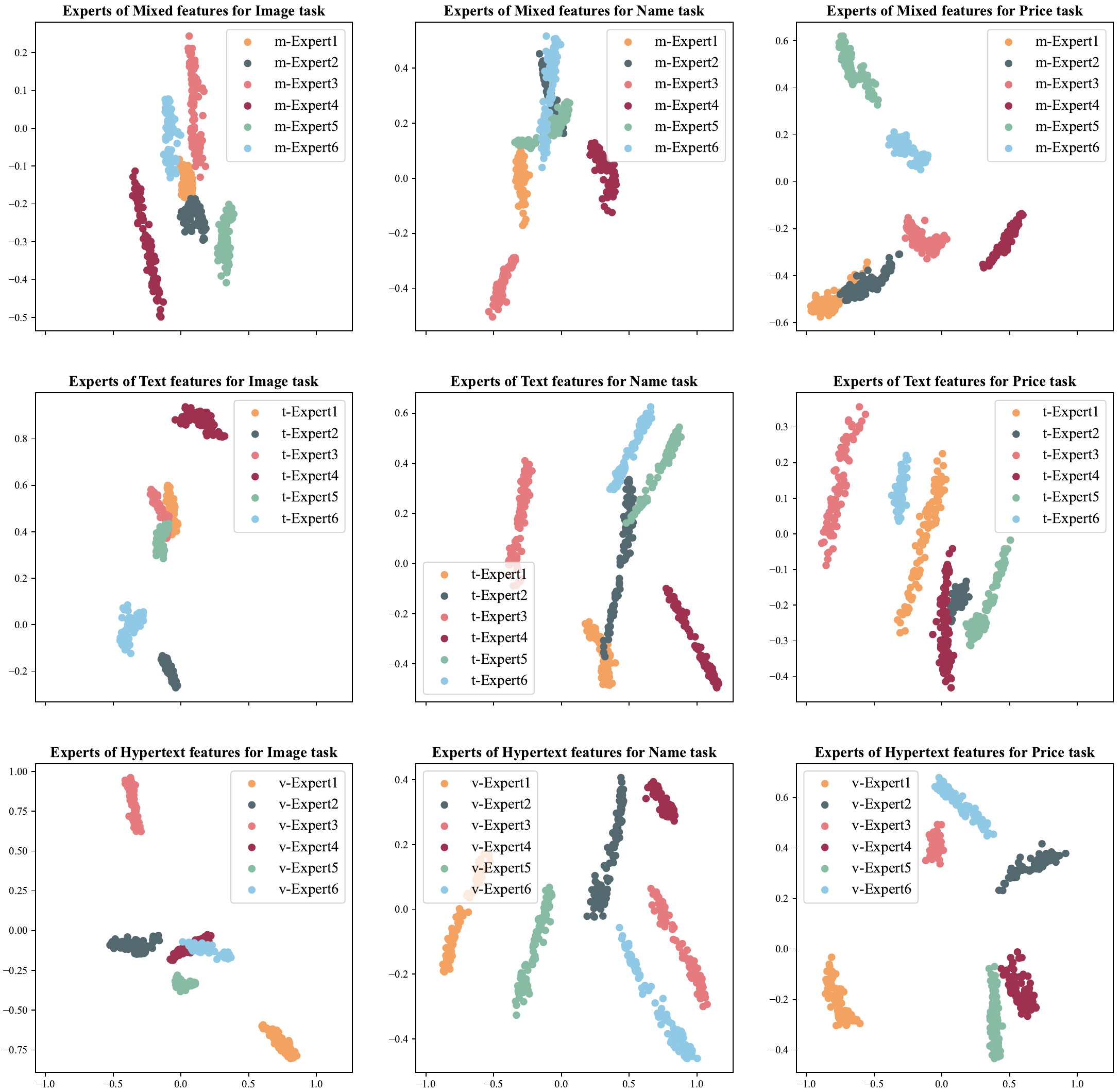}
    \caption{Visualizations of representations for different experts in different tasks. The input representations from different experts naturally form multiple clusters, indicating that the experts from different modalities have differentiation.}
    \label{fig:all_vis_repr}
\end{figure*}

\end{document}